%% file: main.tex
\documentclass[10pt,twocolumn,letterpaper]{article}
\usepackage{stfloats}

\usepackage{cvpr}              
\usepackage[OT1]{fontenc} 


\definecolor{cvprblue}{rgb}{0.21,0.49,0.74}
\usepackage[pagebackref,breaklinks,colorlinks,allcolors=cvprblue]{hyperref}
\def\paperID{2425} 
\def\confName{CVPR}
\def\confYear{2025}
\newcommand\nnfootnote[1]{%
  \begin{NoHyper}
  \renewcommand\thefootnote{}\footnote{#1}%
  \addtocounter{footnote}{-1}%
  \end{NoHyper}
}
\renewcommand{\thefootnote}{}
\hypersetup{
    colorlinks=true,
    urlcolor=purple 
}
\usepackage{cancel}

\title{USP-Gaussian: Unifying Spike-based Image Reconstruction, Pose Correction and Gaussian Splatting}

\author{Kang Chen\textsuperscript{\rm 1,2}\ \ 
    Jiyuan Zhang\textsuperscript{\rm 1,2}\ \
    Zecheng Hao\textsuperscript{\rm 1,2} \ \
    Yajing Zheng\textsuperscript{\rm 1,2} \ \
    Tiejun Huang\textsuperscript{\rm 1,2,3} \ \
    Zhaofei Yu\textsuperscript{\rm 1,3}\footnotemark[1]\\
    \textsuperscript{\rm 1}School of Computer Science, Peking University\\
    \textsuperscript{\rm 2}State Key Laboratory for Multimedia Information Processing, Peking University\\
    \textsuperscript{\rm 3}Institute for Artificial Intelligence, Peking University
    \\
{\tt\small \{mrchenkang,jyzhang\}@stu.pku.edu.cn, \{haozecheng,yj.zheng,tjhuang,yuzf12\}@pku.edu.cn}
}

\usepackage{xcolor} 
\usepackage{graphicx}
\usepackage{caption}
\usepackage{tabularray}
\usepackage{color}
\usepackage{colortbl}
\usepackage{multirow}
\usepackage{mathtools}
\usepackage{cancel}
\usepackage{amssymb} 
\usepackage{pifont}
\usepackage{animate}

\usepackage{booktabs}
\usepackage[accsupp]{axessibility}  

\usepackage{amsmath}
\usepackage[export]{adjustbox}
\newcommand{\MyEmoji}[1]{\includegraphics[width=1em,valign=t]{#1}}
\newcommand{\fire}{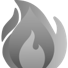}

\begin{document}
\twocolumn[{
\renewcommand\twocolumn[1][]{#1}
\maketitle
\vspace{-3em}
\begin{center}
    \includegraphics[width=1.0\textwidth]{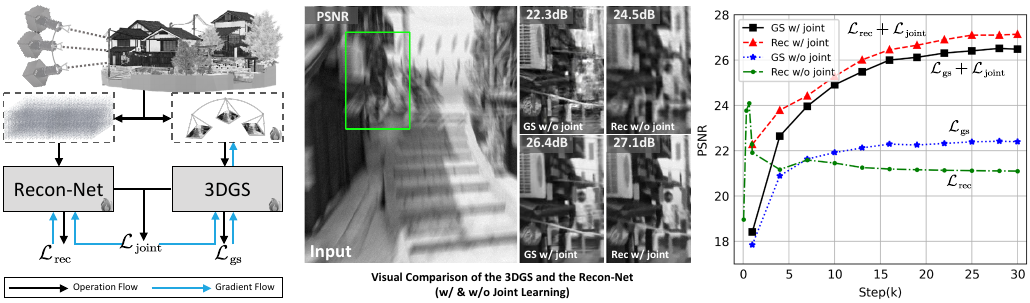}
    \captionof{figure}{\textbf{Left.} Illustration of our USP-Gaussian framework, where the spike-based image Reconstruction Network (Recon-Net), camera poses, and 3DGS are collaboratively optimized signified by \MyEmoji{\fire}. \textbf{Mid.} Visual ablation showcasing the performance of Recon-Net and 3DGS with and without (w/ \& w/o) the joint optimization strategy, with the ablation table depicted in \cref{tab:ablation} and the input formulated in \cref{equ:tfp}. \textbf{Right.} Training curve comparison for Recon-Net and 3DGS with and without joint optimization.}
    \label{fig:top_figure}
\end{center}

}]

\definecolor{blue}{RGB}{33,167,225} 
\definecolor{green}{rgb}{0.3, 0.7, 0.3}   

\maketitle

\nnfootnote{\textsuperscript{$*$} Corresponding author.}

\input{sec/0_abstract}    
\input{sec/1_intro}

\input{sec/2_related}
\input{sec/3_preliminary}

\input{sec/4_methods}

\input{sec/5_exp}

\input{sec/6_conclu}

\section*{Acknowledgments}
We sincerely appreciate Yuyan Chen (HUST) for her valuable suggestions and for polishing the figures. This work was supported by the National Natural Science Foundation of China (62422601, 62176003, 62088102, 62306015), the China Postdoctoral Science Foundation (2023T160015), the Young Elite Scientists Sponsorship Program by CAST (2023QNRC001), the Beijing Municipal Science and Technology Program (Z241100004224004) and the Beijing Nova Program (20230484362,20240484703). 

{
    \small
    \bibliographystyle{ieeenat_fullname}
    \bibliography{main}
}


\end{document}

%% file: sec/0_abstract.tex
\begin{abstract}
Spike camera, as an innovative type of neuromorphic camera that captures scenes with 0-1 bit stream at 40 kHz, is increasingly being employed for the novel view synthesis task building on techniques such as Neural Radiance Fields (NeRF) and 3D Gaussian Splatting (3DGS). Previous spike-based approaches typically follow a three-stage pipeline: I. Spike-to-image reconstruction based on established algorithms. II. Camera poses estimation. III. Novel view synthesis. However, the cascading framework suffers from substantial cumulative errors, i.e., the quality of the initially reconstructed images will impact pose estimation, ultimately limiting the fidelity of the 3D reconstruction. To address this limitation, we propose a synergistic optimization framework \textbf{USP-Gaussian}, which unifies spike-to-image reconstruction, pose correction, and gaussian splatting into an end-to-end pipeline. Leveraging the multi-view consistency afforded by 3DGS and the motion capture capability of the spike camera, our framework enables iterative optimization between the spike-to-image reconstruction network and 3DGS. Experiments on synthetic datasets demonstrate that our method surpasses previous approaches by effectively eliminating cascading errors. Moreover, in real-world scenarios, our method achieves robust 3D reconstruction benefiting from the integration of pose optimization. Our code, data, and trained models are available at \url{https://github.com/chenkang455/USP-Gaussian}.
\end{abstract} 

%% file: sec/1_intro.tex
\section{Introduction}
Spike cameras \cite{spikecamera}, inspired by the human retina \cite{zhu2020retina}, introduce an innovative paradigm for encoding high-speed dynamic scenes. In contrast to conventional imaging systems \cite{exposure} that rely on fixed exposure intervals to integrate visual information, spike cameras continuously sense dynamics by recording the spike stream with an impressive firing rate of 40 kHz. Benefiting from the efficient bit representation of spikes, spike cameras can adeptly detect rapid scenes with significantly diminished bandwidth compared to the high-speed cameras such as the Phantom. 

Capitalizing on the remarkable capability of the spike camera to capture intricate motion and texture details in high-speed environments, recent research \cite{wgse,tfp_tfi,spk2img,STDP_zheng,zhu2020retina,zhao2024boosting,chen2025spikeclip} concentrate on establishing the mapping from bit-encoded spike to visually perceivable images. Subsequent investigations delve into the estimation of optical flow \cite{spike_flow}, facilitation of motion deblur \cite{chen2024spikereveal,spike_deblur,zhang2024deblur} and applications such as high-speed object detection \cite{motion_estimation_zheng}, motion magnification \cite{zhang2024spikemm}, depth estimation \cite{spikedepth} and occlusion removal \cite{spike_sai}.

Neural Radiance Fields \cite{mildenhall2020nerf} and 3D Gaussian Splatting  \cite{3dgs} have emerged as prominent novel view synthesis technologies benefiting from their designed differential training frameworks. Among them, 3DGS employs Gaussian primitives for explicit scene representation and the splatting algorithm for projection, thereby achieving faster rendering speed compared to the pixel-by-pixel approach implemented in NeRF. Building on NeRF and 3DGS, the high-speed and continuous capturing potential of the spike camera can be fully unleashed in 3D reconstruction tasks \cite{zhu2024spikenerf,guo2024spikegs,zhang2024spikegs}.

While the idea of applying the spike camera for 3D reconstruction is trivial and straightforward in a sequential manner, \ie, we can reconstruct images from the spike stream, estimate camera poses and subsequently apply NeRF/3DGS to obtain the 3D representation. Nevertheless, the cascading pipeline introduces significant cumulative errors when the quality of initially recovered images is poor, thereby degrading the accuracy of camera pose estimation and limiting the fidelity of the reconstructed scene.

For example, SpikeGS \cite{zhang2024spikegs} initiates with a self-supervised spike-to-image reconstruction framework built on the Blind Spot Network (BSN) \cite{BSN}, intended to mitigate the performance degradation \cite{bsn_chen,red_chen} encountered by supervised learning methods on real-world scenes. With recovered high-quality images, SpikeGS designs a further constraint from the long spike input and integrates it into the vanilla 3DGS training pipeline. 
While SpikeGS have improved preliminary representation for 3DGS, it remain constrained within the cascading framework thus inheriting the aforementioned cumulative error problem.

To address these challenges, we propose \textit{\textbf{USP-Gaussian}},  \textit{\textbf{U}}nifying \textit{\textbf{S}}pike-based image reconstruction, \textit{\textbf{P}}ose correction, and \textit{\textbf{Gaussian}} Splatting within a one-stage optimization framework to \textit{\textbf{mitigate error propagation inherent in the three-stage cascading pipeline.}}
Specifically, USP-Gaussian comprises two synergistic branches as illustrated in \cref{fig:top_figure}, one dedicated to spike-to-image reconstruction via Recon-Net and the other to novel view synthesis implemented with 3DGS. To facilitate the training of Recon-Net, we design $\mathcal{L}_{\text{rec}}$ based on the constraint between the input spike and recovered outputs. Besides, we supervise the training of 3DGS based on $\mathcal{L}_{\text{gs}}$, which obviates the requirement for pre-recovered high-quality images. While each branch of USP-Gaussian operates independently, we further introduce the joint loss $\mathcal{L}_{\text{joint}}$ derived from the outputs of both two branches to facilitate unified optimization. 

Visual ablation comparisons of restored images and the training curve depicted in \cref{fig:top_figure} demonstrate that \textit{\textbf{the joint optimization of the two branches mutually enhances the recovered textures and the overall training process.}}
We conduct quantitative and qualitative evaluations on the synthetic dataset and the real-world dataset. Experiments showcase that our method achieves superior restoration in both datasets, validating the effectiveness of our joint learning framework. In summary, the contributions of our USP-Gaussian are as follows:
\begin{itemize}
\item[$\bullet$]We propose a unified framework for the joint optimization of Recon-Net, Pose, and 3DGS, mitigating error amplification inherent in cascading frameworks.
\item[$\bullet$] We demonstrate that spike-to-image and 3D reconstruction tasks can mutually facilitate and enhance the optimization of each other.
\item[$\bullet$] We introduce a novel self-supervised training paradigm for the spike-based image reconstruction network with the multi-view consistency afforded by 3DGS.
\end{itemize}

%% file: sec/2_related.tex
\section{Related Work}

\subsection{Spike-based Image Reconstruction}
While neuromorphic cameras generate high frame rate bit-stream information, it is imperative to transform them into multi-bit grayscale images that are perceptible to human observers. To achieve this spike-to-image target, a plethora of methodologies  \cite{wgse,spk2img,STDP_zheng,zhu2020retina,zhao2024boosting,bsn_chen,red_chen,chen2025spikeclip} have been proposed, which can be broadly classified into model-based, supervised, and self-supervised learning paradigms. 
Model-based methods, rooted in the sampling principle of the spike camera, aim to infer high-fidelity images drawing from the physical spike-image relationship  \cite{tfp_tfi}, biological retinal models  \cite{zhu2020retina}, and short-term synaptic plasticity  \cite{STDP_zheng}. 
Confronted with pervasive dark current noise that is challenging to model, learning-based methods are employed to directly deduce the sharp image from spike input with unknown firing patterns. Specifically, Spk2ImgNet \cite{spk2img} designed a hierarchical architecture engineered to capture the intricate temporal dynamics present within spike streams. SpikeCLIP \cite{chen2025spikeclip} utilized the textual description of the captured scene as supervision for training the spike-to-image network. 
Concurrently, self-supervised learning methodologies based on the BSN \cite{bsn_chen,red_chen} are developed to counteract the performance deterioration in supervised learning methods owing to the domain gap between synthetic and real-world datasets. 

In this paper, we propose a novel self-supervised spike-based image reconstruction framework leveraging the multi-view constraints provided by 3D scenes. 

\subsection{Novel View Synthesis based on the Neuromorphic Camera}
Recent research has extensively employed neuromorphic cameras, including event cameras \cite{EventSurvey,li2024benerf,guan1,guan2,guan3}, which encode intensity difference in dynamic scenarios with low-latency and spike cameras \cite{spikecamera}, for 3D scene reconstruction.

\textbf{Event Camera.}  
\citet{E-nerf} proposed the E-NeRF based on the event generation mechanism, which can recover sharp 3D scenes under high-speed camera motion.  
\citet{EventNerf} further harnessed a color event camera to learn the color representation of scenes.  
\cite{evdnerf,event_dnerf} focused on recovering dynamic scenes with rigid deformations from event streams. In addition to NeRF, EvGGS  \cite{wang2024evggs} introduces a generalizable Gaussian splatting framework for event cameras designed to reconstruct 3D scenes without retraining. EF-3DGS  \cite{liao2024ef} develops a free-trajectory framework by aligning the RGB input, event streams, and pose. 
Event3DGS  \cite{xiong2024event3dgs} presents an effective solution for rapid ego-motion robotics scenarios.

\textbf{Spike Camera.}  \citet{zhu2024spikenerf} design a differentiable SNN and integrate it into the NeRF training pipeline to simulate the generation of spike streams, where the loss function is formulated based on the rendered spike output and the input spike.  \citet{dai2024spikenvs} further introduce SpikeNVS by extending the spike cameras to RGB-Spike input. Additionally,  \citet{guo2024spikegs} introduced a color 3D scene reconstruction framework by integrating Bayer-pattern spike streams into the 3DGS training pipeline. SpikeGS \cite{zhang2024spikegs} initially employs the BSN to recover images from the noisy spike stream and subsequently integrates them into the 3DGS.

While the cascaded framework SpikeGS \cite{zhang2024spikegs} introduces propagation errors, we propose a joint optimization framework that concurrently handles spike-based image reconstruction, pose estimation, and novel view synthesis.

%% file: sec/3_preliminary.tex
\section{Preliminary}
\begin{figure}
    \centering
    \includegraphics[width=1\linewidth]{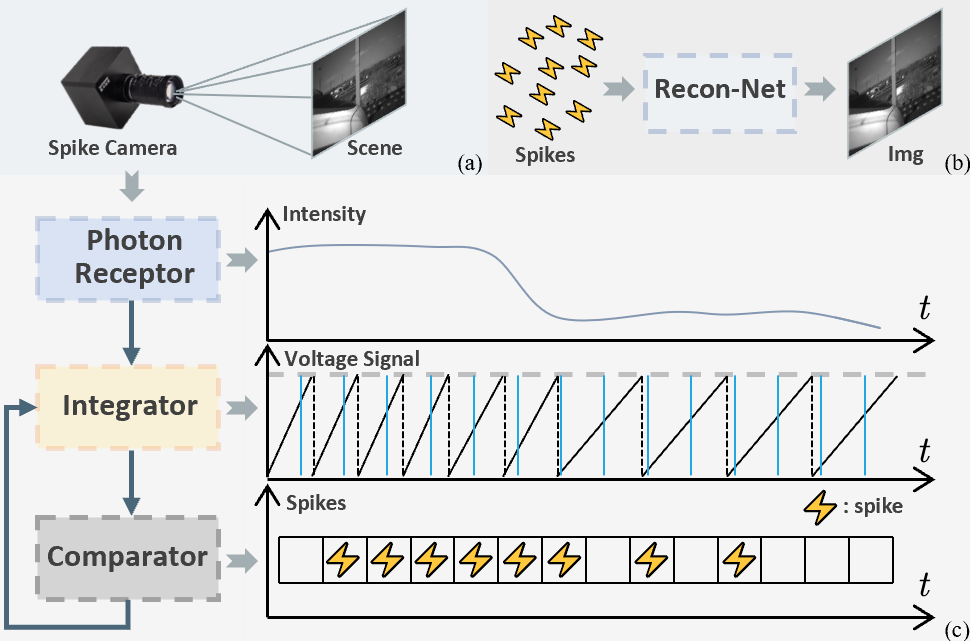}
    \caption{(a) The spike camera captures a railway moving 350 km/h. (b) The Recon-Net is designed to establish the mapping between the input spike stream and the sharp image. (c) The working mechanism of the spike camera.}
    \label{fig:spike_camera}
\end{figure}
\subsection{Spike Camera}
The spike camera consists of the photodetector, voltage integrator, comparator, and synchronous readout as shown in \cref{fig:spike_camera}. Within the designated time interval $\mathcal{T} = [t_s, t_e]$, the photodetector continuously captures the light intensity $\mathbf{I}(t)$ and converts it into the voltage signal $\mathbf{V}(t) = \mu \mathbf{I}(t)$, where $\mu$ is the photoelectric conversion coefficient. Concurrently, the comparator assesses the integrator voltage $\mathbf{A}(t)$ against the predefined threshold $\Theta$. When the threshold is met, the spike is emitted and the integrator voltage is reset, formulated as:
\begin{equation}
    \mathbf{A}(t) = \int_{t_s}^t \mathbf{I}(t)dt \mkern-10mu \mod C, \label{equ:spike_camera}
\end{equation}
where $C = \Theta/\mu$ is a constant proportional to the spike firing threshold $\Theta$. While the integrator continuously aggregates voltage, the readout circuit synchronously retrieves spike signals from the registers at up to 40 kHz. Spikes fired during the specified interval $\mathcal{T}$ are represented as a three-dimensional bit stream $\mathcal{S} \in \{0,1\}^{K \times H \times W}$, where $H$ and $W$ indicate the image height and width, with $K$ denoting the spike sequence length.

\begin{figure*}[t]
    \centering
    \includegraphics[width=1\linewidth]{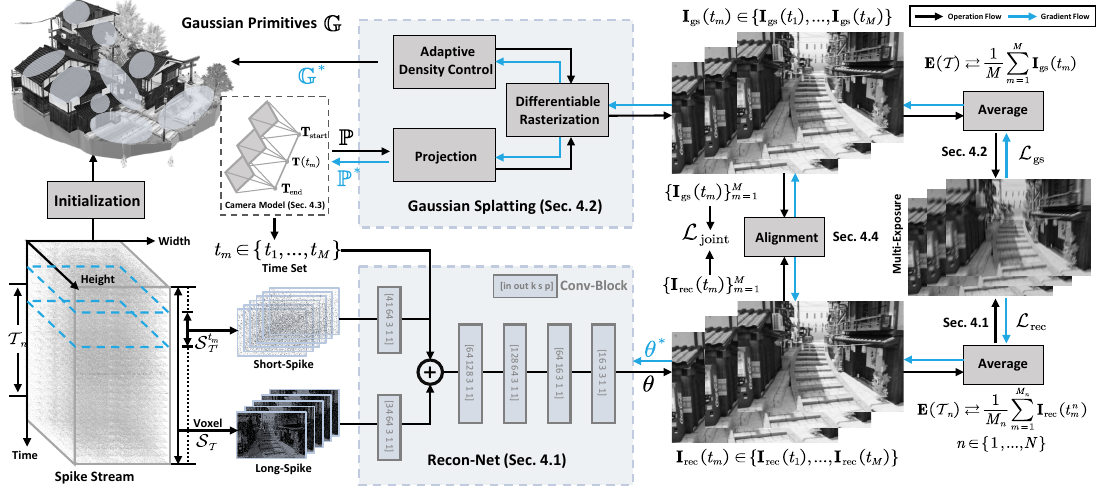}
    \caption{\textbf{The working pipeline of our USP-Gaussian.} 
    For each camera viewpoint, we initially derive a pose sequence at equal time intervals based on the camera model described in \cref{subsec:pose}. Leveraging the 3DGS rendering pipeline detailed in \cref{subsec:spike-gs}, we subsequently generate a corresponding sharp sequence for this viewpoint. Concurrently, Recon-Net is employed to reconstruct the same sequence from the captured spike stream as outlined in \cref{subsec:spike-net}. 
    While Recon-Net and 3DGS are trained based on the discrepancy between the re-synthesized blurry image with the input long-exposure image, alignment of their outputs for the same scene is further enforced through joint optimization loss as described in \cref{subsec:joint}.
    With formulated loss functions, our USP-Gaussian enables the collaborative optimization of Gaussian primitives, camera poses, and the parameters of Recon-Net as formulated in \cref{equ:motivation}.
    }
    \label{fig:framework}
\end{figure*}

\subsection{3D Gaussian Splatting}
3D Gaussian Splatting leverages a set of three-dimensional Gaussian primitives, symbolized as $\{\mathcal{G}\}_{n=1}^{N_{\text{gs}}}$, to effectively represent the spatial characteristics of the captured scene.

Each Gaussian primitive is centered at a point $\mathbf{p}_n$, with its spread and orientation influenced by the covariance matrix $\Sigma_n = RSS^TR^T$, where $R$ and $T$ denotes the rotation and the scale matrix, respectively. The impact of the Gaussian on any point $\mathbf{v}$ in the scene is quantified by:
\begin{equation}
\mathcal{G}_n(\mathbf{v}) = \exp\left(-\frac{1}{2}(\mathbf{v} - \mathbf{p}_n)^T \Sigma_n^{-1} (\mathbf{v} - \mathbf{p}_n)\right).
\end{equation}

When it comes to rendering for viewing transformation $W$, these 3D Gaussian primitives are mapped onto 2D primitives $\mathcal{G}_n^{2D}$, with the covariance matrix $\Sigma_n'$ reformulated as:
\begin{equation}
    \Sigma_n' = JW\Sigma_nW^TJ^T,
\end{equation}
where $J$ represents the Jacobian matrix derived from the affine approximation of the projective transformation.

Further, each pixel on the rendered plane is calculated by rasterizing these 2D Gaussians, sorted by depth, according to the following formulation:
\begin{equation}
C(r) = \sum_{k=1}^{N_{\text{gs}}} T_i \alpha_i \mathbf{c}_i, \\ T_i = \prod_{j=1}^{i-1} (1 - \alpha_j),
\end{equation}
where $D$ represents the indices set reordered by depth, and $D_k$ denotes the index of the Gaussian primitive that is the $k$-th closest to the rendering point.

%% file: sec/4_methods.tex
\section{Methods}
\textbf{Target Clarification.} Given the captured spike stream set $\{\mathcal{S}_v\}_{v=1}^V$ from $V$ different viewpoints with initial inaccurate poses $\{\mathcal{P}_v\}_{v=1}^V$, the USP-Gaussian aims to optimize the parameters of Recon-Net and Gaussian primitives while outputting refined poses as shown in \cref{fig:top_figure}, formulated as:
\begin{equation}
    (\mathbb{S}, \mathbb{P}, \mathbb{G}, \theta) \xrightarrow{\text{USP-Gaussian}} (\textcolor{blue}{\mathbb{P}^*}, \textcolor{blue}{\mathbb{G}^*}, \textcolor{blue}{\theta^*}), \label{equ:motivation}
\end{equation}
where $\mathbb{S}$ and $\mathbb{P}$ represent the sets of spike streams and poses from $M$ viewpoints, $\mathbb{G}$ denotes the ensemble of Gaussian primitives, and $\theta$ encapsulates the Recon-Net parameters. The notation (\textcolor{blue}{*}) indicates the optimized parameters.

\subsection{Spike-based Image Reconstruction Network} \label{subsec:spike-net}
The Recon-Net is designed to facilitate the mapping $\mathcal{F}(\cdot)$ from the spike input to the sharp image as shown in \cref{fig:spike_camera}(b), formulated as $\mathbf{I}(f) = \mathcal{F}(\mathcal{S}_{\mathcal{T}};\theta)$ with $f$ signifying the central moment of the interval $\mathcal{T}$. Prior research \cite{spk2img, wgse, bsn_chen} employs a fixed and brief window (41 frames in their configuration) of the spike stream as the network input. However, it is insufficient for scenarios with low spike firing rates due to the limited information embedded within. 

While extending the window size allows the network to encompass more features, it complicates the feature extraction process and demands large computational resource. To address it, we configure the input of the Recon-Net with a complementary long-short spike stream as shown in \cref{fig:framework}. 

Given a long spike stream $\mathcal{S}_{\mathcal{T}}$, we extract a short spike stream $\mathcal{S}_{\mathcal{T'}}^t \subset \mathcal{S}_{\mathcal{T}}$ centered around time $t$, with the network mapping reformulated as $\mathbf{I}(t) = \mathcal{F}(\mathcal{S}_{\mathcal{T}}, \mathcal{S}_{\mathcal{T'}}^t, t; \theta)$, where time $t$ is incorporated to guide the alignment of the short and long spike. Following the event stream pre-processing in \cite{E-CIR,esl,TRMD}, we apply voxelization to the long spike stream $\mathcal{S}_{\mathcal{T}}$, reducing its dimensions from $K \times H \times W$ to $\frac{K}{4} \times H \times W$ accumulating every $4$ frames. The network architecture consists of basic convolutional blocks as illustrated in \cref{fig:framework} and detailed in the supplementary materials.

According to the spike camera working mechanism described in \cref{equ:spike_camera}, we can reconstruct a long-exposure image $\mathbf{E}(\mathcal{T})$ from the spike input $\mathcal{S}_{\mathcal{T}}$ with the TFP algorithm \cite{tfp_tfi}, mathematically formulated as below:
\begin{equation}
    \mathbf{E}(\mathcal{T}) = \frac{1}{T}\int_{\mathcal{T}} \mathbf{I}(t) \, dt = \frac{C \cdot N}{T}, \label{equ:tfp}
\end{equation}
where $T$ denotes the duration of the interval $\mathcal{T}$, and $N$ is the cumulative number of spikes during this period. 

Motivated by the textural similarity between the blurry image and the long-exposure image, we can employ the motion-reblur loss widely employed in self-supervised motion deblurring frameworks \cite{chen2024spikereveal,evdi,gem,red} to guide the training of the Recon-Net. We initially reconstruct $M$ frames uniformly across the interval to form the image sequence $\{\mathbf{I}_{\text{rec}}(t_m)\}_{m=1}^M$ and further average them to synthesize the reblur image. The loss can be formulated by comparing the reblur image to the long-exposure image, expressed as:
\begin{equation}
    \mathbb{L}_{\text{rec}}(M,\mathcal{T}) = \mathcal{L}\left(\frac{1}{M}\sum_{m=1}^M \mathbf{I}_{\text{rec}}(t_m),\mathbf{E}(\mathcal{T})\right), \label{equ:spike_loss}
\end{equation}
where the loss \(\mathcal{L}\) \cite{3dgs} combines \(L_1\) loss with a D-SSIM term, defined as \(\mathcal{L} = (1-\lambda)\mathcal{L}_1 + \lambda \mathcal{L}_{\text{D-SSIM}}\), effectively integrating both pixel accuracy and structural similarity.

Nevertheless, the input spike stream $\mathcal{S}_{\mathcal{T}}$ sufficiently captures the information to estimate $\mathbf{E}(\mathcal{T})$ as detailed in \cref{equ:tfp}, which will lead the Recon-Net to learn a trivial identity mapping from the spike to the long exposure image. To mitigate it, we propose the multi-reblur loss to regularize the reconstructed sequence. Specifically, we extract $N$ sub-intervals from the long exposure $\mathcal{T}_n \subset \mathcal{T}$ with the same central moment and apply the reblur loss to each sub-interval, formulating the multi-reblur loss as below:
\begin{equation}
\mathcal{L}_{\text{rec}} = \frac{1}{N} \sum_{n=1}^N \mathbb{L}_{\text{rec}}(M_n,\mathcal{T}_n),  \label{equ:spike_loss2}
\end{equation}
where $M_n$ represents the number of reconstructed frames within the $n$-th sub-interval $\mathcal{T}_{n}$.

\subsection{Spike-based Gaussian Splatting} \label{subsec:spike-gs}
Suppose that retrieving the pose for each timestamp within the interval is straightforward. Following \cref{subsec:spike-net}, we uniformly obtain $M$ poses from the interval $\mathcal{T}$, yielding a pose sequence $\{\mathbf{T}(t_m)\}_{m=1}^M$ with the corresponding image sequence $\{\mathbf{I}_{\text{gs}}(t_m)\}_{m=1}^M$ rendered via the 3DGS rasterization pipeline as shown in \cref{fig:framework}.

While reconstructing a sequence of images from the spike stream by shortening the exposure interval in \cref{equ:tfp} for the training of 3DGS sounds feasible, these short-exposure images are noisy and lack texture details \cite{chen2024spikereveal}. To this end, we utilize the long-exposure image $\mathbf{E}(\mathcal{T})$ as the supervision and design the motion-reblur loss for 3DGS similar to \cref{equ:spike_loss}, expressed as:
\begin{equation}
    \mathcal{L}_{\text{gs}} = \mathcal{L}\left(\frac{1}{M}\sum_{m=1}^M \mathbf{I}_{\text{gs}}(t_m) ,\mathbf{E}(\mathcal{T})\right) . \label{equ:3dgs_loss}
\end{equation}

\subsection{Spike Camera Trajectory Modeling} \label{subsec:pose}
Building on the learnable pose representation during camera exposure in previous research \cite{bad-nerf,zhao2024bad}, we employ a similar approach to optimize the camera pose throughout the interval $\mathcal{T}$ as shown in \cref{fig:framework}. For each spike stream $\mathcal{S}_{\mathcal{T}}$, we optimize both the start pose $\mathbf{T}_{\text{start}}$ and end pose $\mathbf{T}_{\text{end}}$ over the interval, while intermediate poses are represented through interpolation between these two.

For the pose $\mathbf{T}(t_m)$ at time $t_m$, we utilize linear interpolation in the Lie algebra of $\text{SE}(3)$ to estimate the pose at time $t$, formulated as:
\begin{equation}
\mathbf{T}(t_m) = \mathbf{T}_{\text{start}} \cdot \exp\left(\frac{t_m}{T} \cdot \log\left( \mathbf{T}_{\text{start}}^{-1} \cdot \mathbf{T}_{\text{end}}\right)\right). \label{equ:pose}
\end{equation}

During the training, the pose set $\mathbb{P}$ consisting of the start and end poses from $V$ viewpoints is jointly optimized with the Recon-Net and Gaussian primitives as in \cref{equ:motivation}. For further details on the camera trajectory modelling, please refer to BAD-NeRF \cite{bad-nerf} and BAD-Gaussian \cite{zhao2024bad}.

\input{imgs_table/img_compare_syn}

\subsection{Joint Optimization} \label{subsec:joint}
Given the spike stream $\mathcal{S}_{\mathcal{T}}$ along with the start pose $\mathbf{T}_{\text{start}}$ and the end pose $\mathbf{T}_{\text{end}}$ to be optimized, we perform evenly spaced sampling interpolation as in \cref{equ:pose} during the interval, obtaining the pose sequence $\{\mathbf{T}(t_m)\}_{m=1}^M$. 

For each pose in the sequence, we select a short spike stream centered around the timestamp $t_m$, obtaining the image sequence $\{\mathbf{I}_{\text{rec}}(t_m)\}_{m=1}^M$ based on the Recon-Net in \cref{subsec:spike-net} and the image sequence $\{\mathbf{I}_{\text{gs}}(t_m)\}_{m=1}^M$ rendered from the 3DGS in \cref{subsec:spike-gs}. While both sequences represent the same scene, the joint optimization loss $\mathcal{L}_{\text{joint}}$ can be formulated as follows:
\begin{equation}
    \mathbb{L}_{\text{joint}} = \frac{1}{M}\sum_{m=1}^M||\mathbf{I}_{\text{gs}}(t_m)-\mathbf{I}_{\text{rec}}(t_m)||_2^2.
\end{equation}

However, we found that the joint optimization framework tends to be unstable due to the order difference between the 3DGS and Recon-Net output sequences. Specifically, since the temporal order does not affect the loss in \cref{equ:3dgs_loss}, the optimized pose sequence may reverse relative to the real camera trajectory. To address this, we apply a simple but effective flip-and-minimum operation to align both sequences, with the loss function reformulated as:
\begin{align}
    \mathcal{L}_{\text{joint}} &= \min(\mathbb{L}_{\text{joint}},\mathbb{L}_{\text{joint}}^{\text{r}}) ,
    \\
    \mathbb{L}_{\text{joint}}^{\text{r}} &= \frac{1}{M}\sum_{m=1}^M||\mathbf{I}_{\text{gs}}(t_{M-m+1})-\mathbf{I}_{\text{rec}}(t_m)||_2^2,
\end{align}
where the difference between the $\mathbb{L}_{\text{joint}}^{\text{r}}$ and the $\mathbb{L}_{\text{joint}}$ lies in that the rendered sequence via 3DGS is reversed.

To sum up, the loss function for our USP-Gaussian framework is presented as follows:
\begin{equation}
    \mathcal{L} = \mathcal{L}_{\text{rec}} +\mathcal{L}_{\text{gs}} + \mathcal{L}_{\text{joint}}.
    \label{equ:final_loss}
\end{equation}
where $\mathcal{L}_{\text{rec}}$ optimizes Recon-Net parameters, $\mathcal{L}_{\text{gs}}$ refines poses and Gaussian primitives, and $\mathcal{L}_{\text{joint}}$ jointly optimizes all components, with gradient flow shown in \cref{fig:framework}.

%% file: imgs_table/img_compare_syn.tex
\begin{figure*}[t]
\centering
\begin{subfigure}{\textwidth}
\centering
\includegraphics[width=1.0\textwidth]{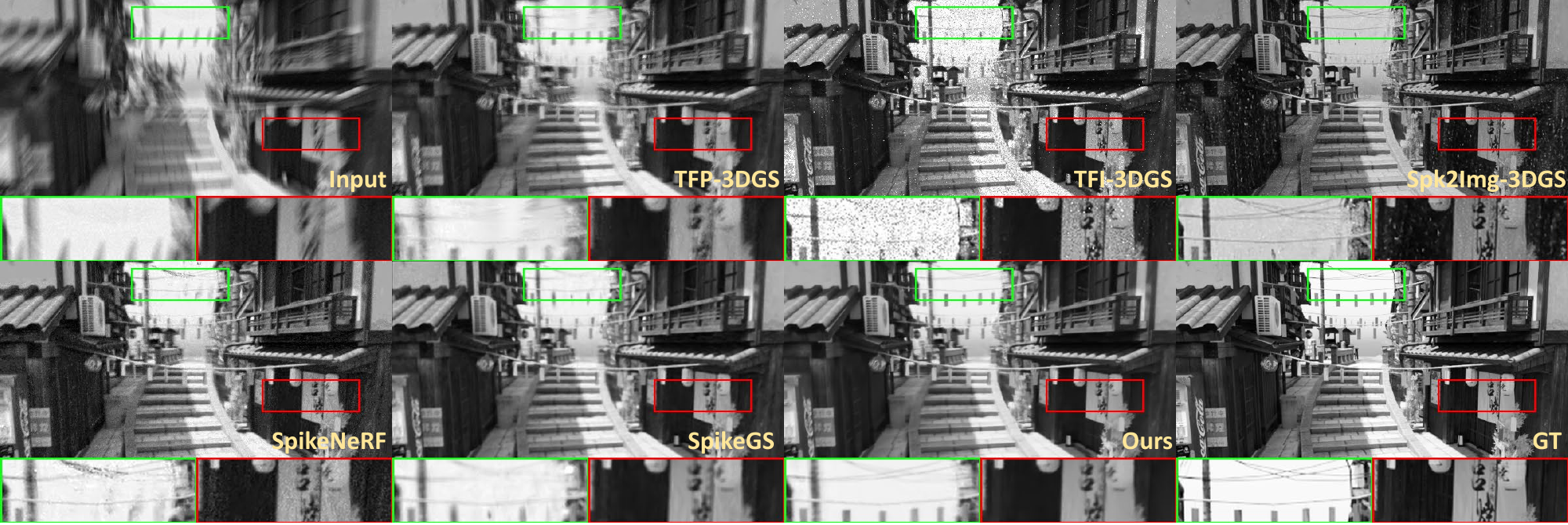}
\vspace{-0.47cm}
\end{subfigure} 
\begin{subfigure}{\textwidth}
\centering
\includegraphics[width=1.0\textwidth]{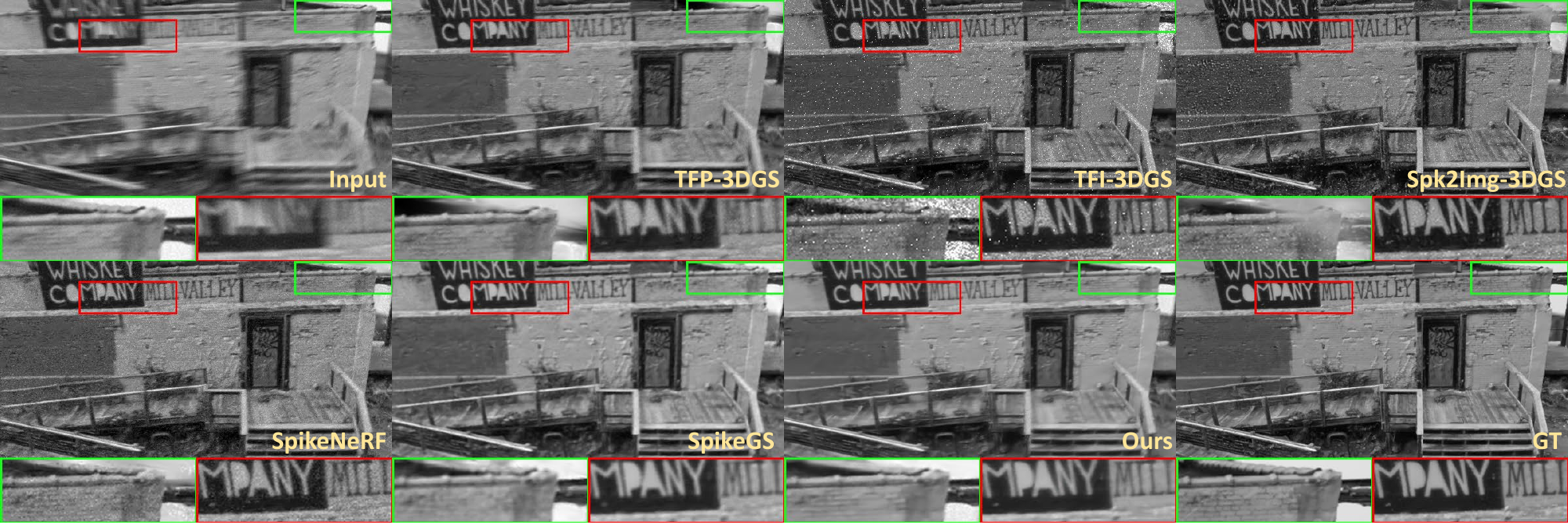}
\end{subfigure}
\caption{3D reconstruction visual comparison of our USP-Gaussian compared with previous methods on the synthetic dataset, where the input is the long-exposure image defined in \cref{equ:tfp}.}
\label{fig:3dgs_compare_syn}
\end{figure*}

%% file: sec/5_exp.tex
\input{imgs_table/table_compare_syn_3dgs}
\input{imgs_table/img_compare_real}

\section{Experiment}
\subsection{Datasets}
\noindent\textbf{Synthetic Dataset.} 
We construct the synthetic dataset based on Deblur-NeRF \cite{deblur_nerf} scenes for quantitative evaluation. We start with rendering $13$ sharp images for each viewpoint, with further interpolation algorithm XVFI \cite{sim2021xvfi} to insert $7$ images between each frame pair. We apply the spike simulator on the interpolated sequence to emulate the real-world spike as in \cite{chen2024spikereveal}, thus generating $97$ spike frames for each viewpoint. 

\noindent\textbf{Real-world Dataset.} 
To construct the real-world dataset, we shake the spike camera extremely fast while maintaining a fixed forward-facing orientation and extract $97$ spike frames per view, consistent with the configuration in the synthetic dataset. 
For pose estimation, we convert the raw spike stream in each view to the short-exposure image as defined in \cref{equ:tfp} and subsequently employ COLMAP \cite{colmap} to estimate the pose sequence. 


\subsection{Experimental Results}
We compare our USP-Gaussian with two-stage cascading pipelines and approaches specially designed for the spike camera, \ie, SpikeNeRF \cite{zhu2024spikenerf} and SpikeGS \cite{zhang2024spikegs}. For the two-stage approach, we initially apply TFP \cite{tfp_tfi}, TFI \cite{tfp_tfi}, and Spk2ImgNet \cite{spk2img} to reconstruct sharp images from the input spike followed by the common 3DGS processing pipeline, with the code implemented by Spike-Zoo \cite{spikezoo}. All Gaussian-based methods are conducted based on the same 3DGS backbone, and identical camera poses input for a fair comparison. While SpikeGS requires several poses per spike stream for the exposure loss, which presents a notable challenge for COLMAP in real-world data, we omit the exposure loss and retain the rest of the framework. All comparative metrics are calculated based on the 3D reconstruction task, with further comparison on the spike-to-image reconstruction task and implementation details available in the supplementary materials.

\input{imgs_table/tab_pose}

\textbf{Synthetic Dataset.} Quantitative and qualitative comparisons on the synthetic dataset are shown in \cref{tab:3dgs_table} and \cref{fig:3dgs_compare_syn} respectively. As illustrated in the \cref{tab:3dgs_table}, our USP-Gaussian achieves the best performance across most metrics in various scenes. Two-stage cascading methods TFP-3DGS and TFI-3DGS yield poor restoration due to the substantial noise embedded in the initially reconstructed sequence. Similarly, although Spk2ImgNet performs well on its original REDS dataset \cite{spk2img}, it experiences substantial degradation due to the dataset domain gap.
SpikeGS, as the primary comparison algorithm in this paper, heavily relies on the BSN \cite{BSN} for self-supervised Spike-to-Image reconstruction. Nevertheless, the denoising performance of the BSN deteriorates in the presence of substantial noise, leading to pronounced grid artifacts and the attenuation of high-frequency details of the SpikeGS. SpikeNeRF faces a comparable challenge, as it requires noisy spike images as input and relies on the SNN for denoising.

To this end, our proposed collaborative learning framework USP-Gaussian, benefiting from the inherent constraints between the spike stream and the multi-view consistency afforded by the 3DGS, effectively addresses the aforementioned problems without relying on the pre-training of the spike-based image reconstruction network. 

In the previous part, accurate poses are obtained by applying COLMAP on sharp images rendered by Blender. To demonstrate that the integration of pose correction enhances the robustness of our method, we apply random perturbations of 10\%, 20\%, and 30\% to the accurate poses in the wine scene, generating inaccurate poses for SpikeGS and our method with the comparison shown in \cref{tab:tab_pose}.

\begin{figure}[t]
    \centering
    \begin{minipage}{0.5\columnwidth}
        \centering
        \includegraphics[width=\columnwidth]{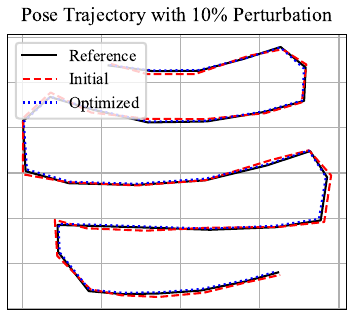} 
    \end{minipage}%
    \hfill
    \begin{minipage}{0.5\columnwidth}
        \centering
        \includegraphics[width=\columnwidth]{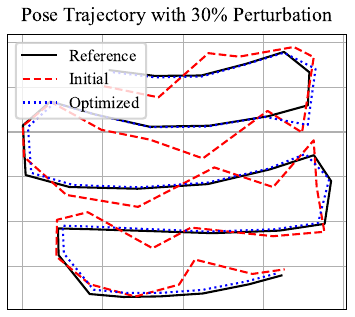}  
    \end{minipage}
    \caption{Visual comparison of initial and optimized poses.}
    \label{fig:pose_compare}
    \vspace{-1em}
\end{figure}

\input{imgs_table/tab_pose_opt}

Furthermore, visual trajectories and quantitative errors of the initial noisy, and optimized poses are depicted in \cref{fig:pose_compare} and \cref{tab:pose_tab_opt} respectively, with MAE measuring translation error and radians measuring rotation error, demonstrating the pose correction capability of our USP-Gaussian.

\textbf{Real-world Dataset.} We also conduct the visual comparison on the real-world dataset as shown in \cref{fig:3dgs_compare_real}. Due to the rapid camera ego-motion and the low imaging quality of the spike camera characterized by low spatial resolution, insufficient texture details, and underexposure, the pose sequence estimated by the COLMAP is highly inaccurate, presenting significant challenges for previous methods to recover accurate 3D representation as observed in \cref{fig:3dgs_compare_real}. Nevertheless, our USP-Gaussian overcomes this limitation benefiting from the unifying optimization of the pose.


\input{imgs_table/tab_abl}
\begin{figure}[t]
    \centering
    \includegraphics[width=1\linewidth]{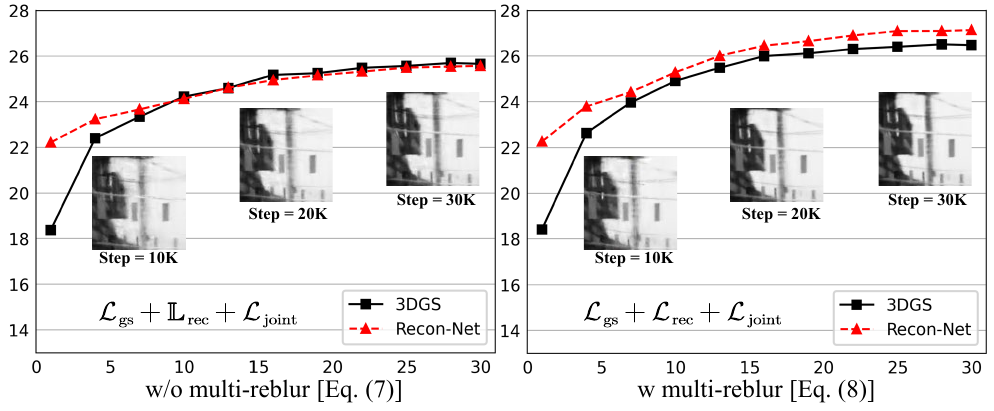}
    \caption{Visual ablation on the effectiveness of multi-reblur loss, with the x-axis and y-axis representing step (k) and PSNR.}
    \vspace{-1em}
    \label{fig:multi_reblur}
\end{figure}

\subsection{Ablation Study}
In this subsection, we demonstrate the effectiveness of our proposed joint optimization, complementary spike input, and multi-reblur loss strategies.

\textbf{I. Joint Learning.} The highlight of USP-Gaussain lies in the design of the joint loss, which aligns outputs from the 3DGS and Recon-Net thus achieving complementary performance improvement. We conduct the ablation experiment on loss functions of $\mathcal{L}_{\text{gs}}$, $\mathcal{L}_{\text{rec}}$ and $\mathcal{L}_{\text{joint}}$ with quantitative results listed in \cref{tab:ablation}, which shows that the incorporation of the joint loss (ID-V) yields notable improvements for both 3DGS and Recon-Net compared to independent optimization (ID-I and ID-III). 

Further visual ablation comparisons of the restored image and the training curve are presented in \cref{fig:top_figure}. Visual comparison substantiates that the joint optimization of 3DGS and Recon-Net significantly enhances the texture details for both two. Regarding the training curve, the independent 3DGS training (ID-I) demonstrates slow convergence due to the absence of crucial information in the long-exposure images. Concurrently, the independent training of Recon-Net (ID-III) tends to overfit due to the simplicity of the spike reconstruction task. To this end, our proposed joint optimization framework provides synergistic benefits for both, \ie, Recon-Net effectively compensates for deficient motion features embedded in the long-exposures image while the multi-view consistency afforded by 3DGS culminates in coherent outputs from the Recon-Net. 

\begin{figure}[t]
    \centering
    \includegraphics[width=1\linewidth]{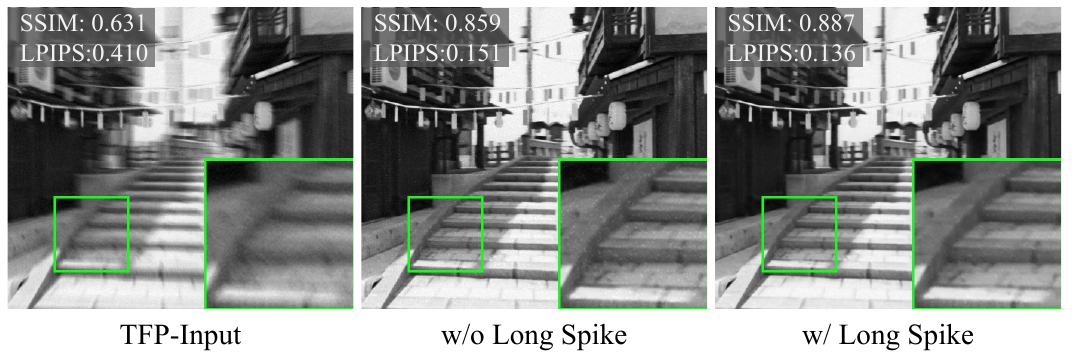}
    \caption{Visual ablation on the effectiveness of long spike input.}
    \vspace{-1em}
    \label{fig:multi_net}
\end{figure}

\textbf{II. Multi-reblur Loss.} The multi-reblur loss $\mathcal{L}_{\text{rec}}$ in \cref{equ:spike_loss} is designed to substitute the single-reblur loss $\mathbb{L}_{\text{rec}}$ in \cref{equ:spike_loss2}. Without further constraints, $\mathbb{L}_{\text{rec}}$ will lead Recon-Net to learn an identity mapping from the input spike stream to the corresponding long-exposure image as demonstrated in \cite{chen2024spikereveal}.  Comparative analysis in \cref{tab:ablation} depicts that  $\mathcal{L}_{\text{rec}}$ (ID-V) yields beneficial improvement than $\mathbb{L}_{\text{rec}}$(ID-IV). Further comparison on the training curve in \cref{fig:multi_reblur} reveals the mechanism of collaborative training, \ie, the multi-reblur loss enhances the performance of Recon-Net, thereby facilitating the training of the 3DGS.

\textbf{III. Complementary Input.} We utilize complementary long and short spike stream inputs for Recon-Net to address the fixed short window limitation inherent in previous methods \cite{spk2img,wgse,bsn_chen}. As illustrated in the visual comparison in \cref{fig:multi_net}, the inclusion of the long spike effectively suppresses salt-and-pepper noise in the Recon-Net output, leading to improved SSIM and LPIPS metrics.

%% file: imgs_table/table_compare_syn_3dgs.tex
\begin{table*}
\caption{3D reconstruction quantitative comparison on the synthetic dataset. Color shading indicates the \colorbox{red!25}{best} and \colorbox{orange!25}{second-best} result.}
\label{tab:3dgs_table}
\centering
\resizebox{\textwidth}{!}{ 
\begin{tabular}{lccc cccc cccc cccc cccc}
\toprule[2pt]
\multirow{2}{*}{Methods} & \multicolumn{3}{c}{Wine} & & \multicolumn{3}{c}{Tanabata} & & \multicolumn{3}{c}{Factory} & & \multicolumn{3}{c}{Outdoorpool} & & \multicolumn{3}{c}{Avg} \\
\cmidrule(lr){2-4} \cmidrule(lr){6-8} \cmidrule(lr){10-12} \cmidrule(lr){14-16} \cmidrule(lr){18-20}
                        & PSNR & SSIM & LPIPS & & PSNR & SSIM & LPIPS & & PSNR & SSIM & LPIPS & & PSNR & SSIM & LPIPS & & PSNR & SSIM & LPIPS \\
\midrule
TFP-3DGS \cite{tfp_tfi}       & 24.138 & 0.796 & 0.239 & & 24.626 & 0.778 & 0.274 & & 27.845 & 0.852 & \cellcolor{orange!25}0.179 & & 28.024 & 0.786 & 0.294 & & 26.158 & 0.803 & 0.239 \\
TFI-3DGS \cite{tfp_tfi}       & 23.846 & 0.733 & 0.251 & & 21.168 & 0.550 & 0.529 & & 25.019 & 0.734 & 0.278 & & 24.755 & 0.637 & 0.446 & & 23.697 & 0.663 & 0.376 \\
Spk2Img-3DGS \cite{spk2img}   & 21.150 & 0.789 & \cellcolor{orange!25}0.178 & & 20.451 & 0.758 & \cellcolor{orange!25}0.214 & & 25.115 & 0.806 & \cellcolor{red!25}0.143 & & 23.362 & 0.647 & 0.417 & & 22.520 & 0.750 & \cellcolor{orange!25}0.238 \\
SpikeNeRF \cite{zhu2024spikenerf} & \cellcolor{orange!25}25.888 & 0.798 & 0.200 & & 25.965 & 0.725 & 0.295 & & 27.531 & 0.780 & 0.191 & & 25.642 & 0.657 & 0.396 & & 26.257 & 0.740 & 0.270 \\
SpikeGS \cite{zhang2024spikegs}   & 25.698 & \cellcolor{orange!25}0.835 & 0.247 & & \cellcolor{orange!25}26.174 & \cellcolor{orange!25}0.826 & 0.243 & & \cellcolor{orange!25}28.627 & \cellcolor{red!25}0.864 & 0.214 & & 28.287\cellcolor{orange!25} & \cellcolor{orange!25}0.797 & \cellcolor{orange!25}0.273 & & \cellcolor{orange!25}27.196 & \cellcolor{orange!25}0.832 & 0.244 \\
Ours      &  \cellcolor{red!25}26.476 & \cellcolor{red!25}0.862 & \cellcolor{red!25}0.176 & & \cellcolor{red!25}26.300 & \cellcolor{red!25}0.827 & \cellcolor{red!25}0.213 & & \cellcolor{red!25}28.694 & \cellcolor{orange!25}0.861 & 0.207 & & \cellcolor{red!25}30.142 & \cellcolor{red!25}0.820 & \cellcolor{red!25}0.270 & & \cellcolor{red!25}27.903 & \cellcolor{red!25}0.843 & \cellcolor{red!25}0.217 \\
\bottomrule[2pt]
\end{tabular}
}
\end{table*}

%% file: imgs_table/img_compare_real.tex
\begin{figure*}[t]
    \centering
    \includegraphics[width=1\linewidth]{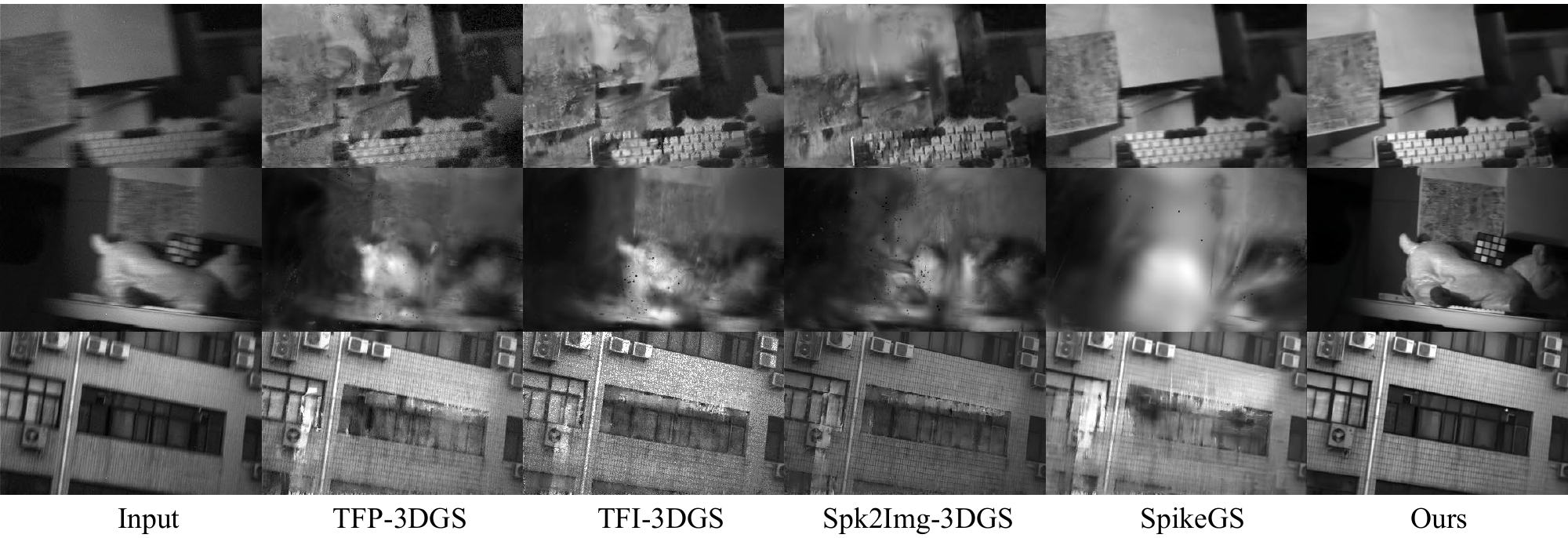}
    \caption{3D reconstruction visual comparison of our USP-Gaussian compared with previous methods on the real-world dataset.}
    \label{fig:3dgs_compare_real}
\end{figure*}

%% file: imgs_table/tab_pose.tex
\begin{table}[t]
\centering
\caption{Comparison of our method with SpikeGS on the wine dataset with inaccurate perturbed poses.}
\label{tab:tab_pose}
\resizebox{\columnwidth}{!}{
\begin{tabular}{lccccccccc}
\toprule[2pt]
   \multirow{2}{*}{Metrics}   & \multicolumn{4}{c}{SpikeGS}  &  & \multicolumn{4}{c}{Ours}  \\
\cmidrule{2-5} \cmidrule{7-10}
 & 0\% & 10\% & 20\% & 30\% &  & 0\% & 10\% & 20\% & 30\%  \\
\midrule
PSNR $\uparrow$  & 25.69 & 18.83 & 17.75 & 16.44 & & 26.47 & 26.05 & 25.93 & 23.46 \\
SSIM $\uparrow$ & 0.835  & 0.607  & 0.592  & 0.567  & & 0.862  & 0.854  & 0.851  & 0.772 \\
LPIPS $\downarrow$ & 0.247  & 0.372  & 0.411  & 0.446  & & 0.176  & 0.184  & 0.186  & 0.245 \\
\bottomrule[2pt]
\end{tabular}}
\vspace{-1.5em}
\end{table}

%% file: imgs_table/tab_pose_opt.tex
\begin{table}[t]
\caption{Translation and rotation poses error comparison.}
\label{tab:pose_tab_opt}
\centering
\resizebox{\columnwidth}{!}{
\begin{tabular}{lccccccccc} 
\toprule[2pt]
\multirow{2}{*}{State}  & \multicolumn{4}{c}{Translation $\downarrow$}  & & \multicolumn{4}{c}{Rotation $\downarrow$} \\ 
\cmidrule{2-5} \cmidrule{7-10} 
     & 10\%  & 20\%  & 30\% & \textbf{Avg} &  & 10\%  & 20\%  & 30\% & \textbf{Avg} \\ 
\midrule
Initial     & 1.302  & 2.967  & 4.193 & 2.820 & &  1.132  & 2.150  & 3.159 & 2.147 \\
Optimized  & 0.855  & 1.781  & 1.493 & 1.376 & & 0.174  & 0.320  & 1.285 & 0.593 \\
\textcolor{green}{$\Delta$ (\%)} & \textcolor{green}{\textbf{34.3\%}} & \textcolor{green}{\textbf{39.9\%}} & \textcolor{green}{\textbf{64.4\%}} & \textcolor{green}{\textbf{51.3\%}} & & \textcolor{green}{\textbf{84.6\%}} & \textcolor{green}{\textbf{85.1\%}} & \textcolor{green}{\textbf{59.2\%}} & \textcolor{green}{\textbf{72.4\%}}  \\
\bottomrule[2pt]
\end{tabular}}
\vspace{-1.5em}
\end{table}

%% file: imgs_table/tab_abl.tex
\begin{table}
\caption{Quantitative ablation on the synthetic dataset.}
\label{tab:ablation}
\centering
\resizebox{0.48\textwidth}{!}{ 
\begin{tabular}{cccccccccccc} 
\toprule[2pt]
\multirow{2}{*}{ID} & \multirow{2}{*}{$\mathcal{L}_{\text{gs}}$} & \multicolumn{2}{c}{Recon-Net} & \multirow{2}{*}{$\mathcal{L}_{\text{joint}}$} & \multicolumn{3}{c}{3DGS Outputs} & & \multicolumn{3}{c}{Recon-Net Outputs} \\ 
\cmidrule{3-4}\cmidrule{6-8}\cmidrule{10-12}
                    &                     & $\mathbb{L}_{\text{rec}}$ & $\mathcal{L}_{\text{rec}}$ & & PSNR & SSIM & LPIPS & & PSNR & SSIM & LPIPS \\ 
\midrule
I                   & $\checkmark$                  &       &                       &                        & 22.392           & 0.750             & 0.208            & & \textbackslash{} & \textbackslash{} & \textbackslash{}  \\
II                  &                    & $\checkmark$    &                       &                        & \textbackslash{} & \textbackslash{} & \textbackslash{} & & 24.255           & 0.708            & 0.358             \\
III                 &                    &     & $\checkmark$                    &                        & \textbackslash{} & \textbackslash{} & \textbackslash{} & & 24.567           & 0.744            & 0.332             \\
IV                  & $\checkmark$                  & $\checkmark$    &                       & $\checkmark$                     & \cellcolor{orange!25}25.659           & \cellcolor{orange!25}0.850             & \cellcolor{orange!25}0.184            & & \cellcolor{orange!25}25.571          & \cellcolor{orange!25}0.807            & \cellcolor{orange!25}0.244             \\
V                   & $\checkmark$                  &     & $\checkmark$                    & $\checkmark$                     & \cellcolor{red!25}26.476           & \cellcolor{red!25}0.862            & \cellcolor{red!25}0.176            & & \cellcolor{red!25}27.138           &\cellcolor{red!25}0.855            &\cellcolor{red!25}0.178             \\
\bottomrule[2pt]
\end{tabular}
}
\end{table}

%% file: sec/6_conclu.tex
\section{Conclusion}
In this paper, we propose a unifying framework USP-Gaussian for the collaborative optimization of spike-based image reconstruction, pose correction, and Gaussian Splatting. By mitigating error propagation inherent in traditional cascading frameworks, USP-Gaussian attains high reconstruction fidelity across both synthetic and real-world datasets, with further ablation experiments revealing that Spike-to-Image reconstruction and 3D reconstruction tasks furnish mutually complementary information.

\textbf{Limitation.} The limitation of USP-Gaussian is that it requires longer training time and more GPU memory compared to previous methods due to the additional poses and Recon-Net parameters to be optimized.

%% file: main.bbl
\begin{thebibliography}{50}
\providecommand{\natexlab}[1]{#1}
\providecommand{\url}[1]{\texttt{#1}}
\expandafter\ifx\csname urlstyle\endcsname\relax
  \providecommand{\doi}[1]{doi: #1}\else
  \providecommand{\doi}{doi: \begingroup \urlstyle{rm}\Url}\fi

\bibitem[Bhattacharya et~al.(2024)Bhattacharya, Madaan, Cladera, Vemprala, Bonatti, Daniilidis, Kapoor, Kumar, Matni, and Gupta]{evdnerf}
Anish Bhattacharya, Ratnesh Madaan, Fernando Cladera, Sai Vemprala, Rogerio Bonatti, Kostas Daniilidis, Ashish Kapoor, Vijay Kumar, Nikolai Matni, and Jayesh~K Gupta.
\newblock Evdnerf: Reconstructing event data with dynamic neural radiance fields.
\newblock In \emph{WACV}, pages 5846--5855, 2024.

\bibitem[Chen and Yu(2024)]{TRMD}
Kang Chen and Lei Yu.
\newblock Motion deblur by learning residual from events.
\newblock \emph{IEEE TMM}, 2024.

\bibitem[Chen et~al.(2025{\natexlab{a}})Chen, Chen, Zhang, Zhang, Zheng, Huang, and Yu]{chen2024spikereveal}
Kang Chen, Shiyan Chen, Jiyuan Zhang, Baoyue Zhang, Yajing Zheng, Tiejun Huang, and Zhaofei Yu.
\newblock Spikereveal: Unlocking temporal sequences from real blurry inputs with spike streams.
\newblock \emph{Advances in Neural Information Processing Systems}, 37:\penalty0 62673--62696, 2025{\natexlab{a}}.

\bibitem[Chen et~al.(2025{\natexlab{b}})Chen, Ye, Huang, and Yu]{spikezoo}
Kang Chen, Zhiyuan Ye, Tiejun Huang, and Zhaofei Yu.
\newblock {Spike-Zoo}: A toolbox for spike-to-image reconstruction.
\newblock \url{https://github.com/chenkang455/Spike-Zoo}, 2025{\natexlab{b}}.

\bibitem[Chen et~al.(2025{\natexlab{c}})Chen, Zheng, Huang, and Yu]{chen2025spikeclip}
Kang Chen, Yajing Zheng, Tiejun Huang, and Zhaofei Yu.
\newblock Rethinking high-speed image reconstruction framework with spike camera.
\newblock \emph{arXiv preprint arXiv:2501.04477}, 2025{\natexlab{c}}.

\bibitem[Chen et~al.(2022)Chen, Duan, Yu, Xiong, and Huang]{bsn_chen}
Shiyan Chen, Chaoteng Duan, Zhaofei Yu, Ruiqin Xiong, and Tiejun Huang.
\newblock Self-supervised mutual learning for dynamic scene reconstruction of spiking camera.
\newblock In \emph{IJCAI}, pages 2859--2866, 2022.

\bibitem[Chen et~al.(2023{\natexlab{a}})Chen, Yu, and Huang]{red_chen}
Shiyan Chen, Zhaofei Yu, and Tiejun Huang.
\newblock Self-supervised joint dynamic scene reconstruction and optical flow estimation for spiking camera.
\newblock In \emph{AAAI}, pages 350--358, 2023{\natexlab{a}}.

\bibitem[Chen et~al.(2023{\natexlab{b}})Chen, Zhang, Zheng, Huang, and Yu]{spike_deblur}
Shiyan Chen, Jiyuan Zhang, Yajing Zheng, Tiejun Huang, and Zhaofei Yu.
\newblock Enhancing motion deblurring in high-speed scenes with spike streams.
\newblock In \emph{NeurIPS}, 2023{\natexlab{b}}.

\bibitem[Dai et~al.(2024)Dai, Wang, Xu, Cheng, Lu, Shi, Zhang, and Huang]{dai2024spikenvs}
Gaole Dai, Zhenyu Wang, Qinwen Xu, Wen Cheng, Ming Lu, Boxing Shi, Shanghang Zhang, and Tiejun Huang.
\newblock Spikenvs: Enhancing novel view synthesis from blurry images via spike camera.
\newblock \emph{arXiv preprint arXiv:2404.06710}, 2024.

\bibitem[Gallego et~al.(2020)Gallego, Delbr{\"u}ck, Orchard, Bartolozzi, Taba, Censi, Leutenegger, Davison, Conradt, Daniilidis, et~al.]{EventSurvey}
Guillermo Gallego, Tobi Delbr{\"u}ck, Garrick Orchard, Chiara Bartolozzi, Brian Taba, Andrea Censi, Stefan Leutenegger, Andrew~J Davison, J{\"o}rg Conradt, Kostas Daniilidis, et~al.
\newblock Event-based vision: A survey.
\newblock \emph{IEEE TPAMI}, 44\penalty0 (1):\penalty0 154--180, 2020.

\bibitem[Guo et~al.(2024)Guo, Hu, Ma, and Huang]{guo2024spikegs}
Yijia Guo, Liwen Hu, Lei Ma, and Tiejun Huang.
\newblock Spikegs: Reconstruct 3d scene via fast-moving bio-inspired sensors.
\newblock \emph{arXiv preprint arXiv:2407.03771}, 2024.

\bibitem[Huang et~al.(2023)Huang, Zheng, Yu, Chen, Li, Xiong, Ma, Zhao, Dong, Zhu, et~al.]{spikecamera}
Tiejun Huang, Yajing Zheng, Zhaofei Yu, Rui Chen, Yuan Li, Ruiqin Xiong, Lei Ma, Junwei Zhao, Siwei Dong, Lin Zhu, et~al.
\newblock 1000$\times$ faster camera and machine vision with ordinary devices.
\newblock \emph{Engineering}, 25:\penalty0 110--119, 2023.

\bibitem[Kerbl et~al.(2023)Kerbl, Kopanas, Leimk{\"u}hler, and Drettakis]{3dgs}
Bernhard Kerbl, Georgios Kopanas, Thomas Leimk{\"u}hler, and George Drettakis.
\newblock 3d gaussian splatting for real-time radiance field rendering.
\newblock \emph{ACM Trans. Graph.}, 42\penalty0 (4):\penalty0 139--1, 2023.

\bibitem[Klenk et~al.(2023)Klenk, Koestler, Scaramuzza, and Cremers]{E-nerf}
Simon Klenk, Lukas Koestler, Davide Scaramuzza, and Daniel Cremers.
\newblock E-nerf: Neural radiance fields from a moving event camera.
\newblock \emph{IEEE Robotics and Automation Letters}, 8\penalty0 (3):\penalty0 1587--1594, 2023.

\bibitem[Laine et~al.(2019)Laine, Karras, Lehtinen, and Aila]{BSN}
Samuli Laine, Tero Karras, Jaakko Lehtinen, and Timo Aila.
\newblock High-quality self-supervised deep image denoising.
\newblock \emph{NeurIPS}, 32, 2019.

\bibitem[Li et~al.(2024)Li, Wan, Wang, Li, Zhou, and Liu]{li2024benerf}
Wenpu Li, Pian Wan, Peng Wang, Jinghang Li, Yi Zhou, and Peidong Liu.
\newblock Benerf: Neural radiance fields from a single blurry image and event stream.
\newblock In \emph{European Conference on Computer Vision (ECCV)}, 2024.

\bibitem[Liao et~al.(2024)Liao, Zhai, Wan, Zhang, Cao, and Zha]{liao2024ef}
Bohao Liao, Wei Zhai, Zengyu Wan, Tianzhu Zhang, Yang Cao, and Zheng-Jun Zha.
\newblock Ef-3dgs: Event-aided free-trajectory 3d gaussian splatting.
\newblock \emph{arXiv preprint arXiv:2410.15392}, 2024.

\bibitem[Liu et~al.(2024)Liu, Guan, Shang, Liang, Yu, and Yu]{guan1}
Zibin Liu, Banglei Guan, Yang Shang, Shunkun Liang, Zhenbao Yu, and Qifeng Yu.
\newblock Optical flow-guided 6dof object pose tracking with an event camera.
\newblock In \emph{Proceedings of the 32nd ACM International Conference on Multimedia}, pages 6501--6509, 2024.

\bibitem[Ma et~al.(2022)Ma, Li, Liao, Zhang, Wang, Wang, and Sander]{deblur_nerf}
Li Ma, Xiaoyu Li, Jing Liao, Qi Zhang, Xuan Wang, Jue Wang, and Pedro~V Sander.
\newblock Deblur-nerf: Neural radiance fields from blurry images.
\newblock In \emph{CVPR}, pages 12861--12870, 2022.

\bibitem[Ma et~al.(2023)Ma, Paudel, Chhatkuli, and Van~Gool]{event_dnerf}
Qi Ma, Danda~Pani Paudel, Ajad Chhatkuli, and Luc Van~Gool.
\newblock Deformable neural radiance fields using rgb and event cameras.
\newblock In \emph{ICCV}, pages 3590--3600, 2023.

\bibitem[Mildenhall et~al.(2020)Mildenhall, Srinivasan, Tancik, Barron, Ramamoorthi, and Ng]{mildenhall2020nerf}
Ben Mildenhall, Pratul~P. Srinivasan, Matthew Tancik, Jonathan~T. Barron, Ravi Ramamoorthi, and Ren Ng.
\newblock Nerf: Representing scenes as neural radiance fields for view synthesis.
\newblock In \emph{ECCV}, 2020.

\bibitem[Peterson(2016)]{exposure}
Bryan Peterson.
\newblock \emph{Understanding exposure: how to shoot great photographs with any camera}.
\newblock AmPhoto books, 2016.

\bibitem[Rudnev et~al.(2023)Rudnev, Elgharib, Theobalt, and Golyanik]{EventNerf}
Viktor Rudnev, Mohamed Elgharib, Christian Theobalt, and Vladislav Golyanik.
\newblock Eventnerf: Neural radiance fields from a single colour event camera.
\newblock In \emph{CVPR}, pages 4992--5002, 2023.

\bibitem[Schonberger and Frahm(2016)]{colmap}
Johannes~L Schonberger and Jan-Michael Frahm.
\newblock Structure-from-motion revisited.
\newblock In \emph{CVPR}, pages 4104--4113, 2016.

\bibitem[Sim et~al.(2021)Sim, Oh, and Kim]{sim2021xvfi}
Hyeonjun Sim, Jihyong Oh, and Munchurl Kim.
\newblock Xvfi: extreme video frame interpolation.
\newblock In \emph{ICCV}, pages 14489--14498, 2021.

\bibitem[Song et~al.(2022)Song, Huang, and Bajaj]{E-CIR}
Chen Song, Qixing Huang, and Chandrajit Bajaj.
\newblock E-cir: Event-enhanced continuous intensity recovery.
\newblock In \emph{CVPR}, pages 7803--7812, 2022.

\bibitem[Tao et~al.(2024)Tao, Li, Guan, Shang, and Yu]{guan2}
Jing Tao, You Li, Banglei Guan, Yang Shang, and Qifeng Yu.
\newblock Simultaneous enhancement and noise suppression under complex illumination conditions.
\newblock \emph{IEEE Transactions on Instrumentation and Measurement}, 2024.

\bibitem[Wang et~al.(2020)Wang, He, Yu, Xia, and Yang]{esl}
Bishan Wang, Jingwei He, Lei Yu, Gui-Song Xia, and Wen Yang.
\newblock Event enhanced high-quality image recovery.
\newblock In \emph{ECCV}, pages 155--171. Springer, 2020.

\bibitem[Wang et~al.(2024)Wang, He, Zhang, Sun, Sun, and Xu]{wang2024evggs}
Jiaxu Wang, Junhao He, Ziyi Zhang, Mingyuan Sun, Jingkai Sun, and Renjing Xu.
\newblock Evggs: A collaborative learning framework for event-based generalizable gaussian splatting.
\newblock \emph{arXiv preprint arXiv:2405.14959}, 2024.

\bibitem[Wang et~al.(2023)Wang, Zhao, Ma, and Liu]{bad-nerf}
Peng Wang, Lingzhe Zhao, Ruijie Ma, and Peidong Liu.
\newblock Bad-nerf: Bundle adjusted deblur neural radiance fields.
\newblock In \emph{CVPR}, pages 4170--4179, 2023.

\bibitem[Xiong et~al.(2024)Xiong, Wu, He, Fermuller, Aloimonos, Huang, and Metzler]{xiong2024event3dgs}
Tianyi Xiong, Jiayi Wu, Botao He, Cornelia Fermuller, Yiannis Aloimonos, Heng Huang, and Christopher Metzler.
\newblock Event3dgs: Event-based 3d gaussian splatting for high-speed robot egomotion.
\newblock In \emph{8th Annual Conference on Robot Learning}, 2024.

\bibitem[Xu et~al.(2021)Xu, Yu, Wang, Yang, Xia, Jia, Qiao, and Liu]{red}
Fang Xu, Lei Yu, Bishan Wang, Wen Yang, Gui-Song Xia, Xu Jia, Zhendong Qiao, and Jianzhuang Liu.
\newblock Motion deblurring with real events.
\newblock In \emph{ICCV}, pages 2583--2592, 2021.

\bibitem[Zhang et~al.(2024{\natexlab{a}})Zhang, Zheng, Chen, Zhang, Chen, Yu, and Huang]{zhang2024spikemm}
Baoyue Zhang, Yajing Zheng, Shiyan Chen, Jiyuan Zhang, Kang Chen, Zhaofei Yu, and Tiejun Huang.
\newblock Spikemm: Flexi-magnification of high-speed micro-motions.
\newblock \emph{arXiv preprint arXiv:2406.00383}, 2024{\natexlab{a}}.

\bibitem[Zhang et~al.(2022)Zhang, Tang, Yu, Lu, and Huang]{spikedepth}
Jiyuan Zhang, Lulu Tang, Zhaofei Yu, Jiwen Lu, and Tiejun Huang.
\newblock Spike transformer: Monocular depth estimation for spiking camera.
\newblock In \emph{ECCV}, pages 34--52. Springer, 2022.

\bibitem[Zhang et~al.(2023{\natexlab{a}})Zhang, Chen, Zheng, Yu, and Huang]{spike_sai}
Jiyuan Zhang, Shiyan Chen, Yajing Zheng, Zhaofei Yu, and Tiejun Huang.
\newblock Unveiling the potential of spike streams for foreground occlusion removal from densely continuous views.
\newblock \emph{arXiv preprint arXiv:2307.00821}, 2023{\natexlab{a}}.

\bibitem[Zhang et~al.(2023{\natexlab{b}})Zhang, Jia, Yu, and Huang]{wgse}
Jiyuan Zhang, Shanshan Jia, Zhaofei Yu, and Tiejun Huang.
\newblock Learning temporal-ordered representation for spike streams based on discrete wavelet transforms.
\newblock In \emph{AAAI}, pages 137--147, 2023{\natexlab{b}}.

\bibitem[Zhang et~al.(2024{\natexlab{b}})Zhang, Chen, Chen, Zheng, Huang, and Yu]{zhang2024spikegs}
Jiyuan Zhang, Kang Chen, Shiyan Chen, Yajing Zheng, Tiejun Huang, and Zhaofei Yu.
\newblock Spikegs: 3d gaussian splatting from spike streams with high-speed camera motion.
\newblock \emph{arXiv preprint arXiv:2407.10062}, 2024{\natexlab{b}}.

\bibitem[Zhang et~al.(2024{\natexlab{c}})Zhang, Chen, Zheng, Yu, and Huang]{zhang2024deblur}
Jiyuan Zhang, Shiyan Chen, Yajing Zheng, Zhaofei Yu, and Tiejun Huang.
\newblock Spike-guided motion deblurring with unknown modal spatiotemporal alignment.
\newblock In \emph{CVPR}, pages 25047--25057, 2024{\natexlab{c}}.

\bibitem[Zhang and Yu(2022)]{evdi}
Xiang Zhang and Lei Yu.
\newblock Unifying motion deblurring and frame interpolation with events.
\newblock In \emph{CVPR}, pages 17765--17774, 2022.

\bibitem[Zhang et~al.(2023{\natexlab{c}})Zhang, Yu, Yang, Liu, and Xia]{gem}
Xiang Zhang, Lei Yu, Wen Yang, Jianzhuang Liu, and Gui-Song Xia.
\newblock Generalizing event-based motion deblurring in real-world scenarios.
\newblock In \emph{ICCV}, pages 10734--10744, 2023{\natexlab{c}}.

\bibitem[Zhao et~al.(2021)Zhao, Xiong, Liu, Zhang, and Huang]{spk2img}
Jing Zhao, Ruiqin Xiong, Hangfan Liu, Jian Zhang, and Tiejun Huang.
\newblock Spk2imgnet: Learning to reconstruct dynamic scene from continuous spike stream.
\newblock In \emph{CVPR}, pages 11996--12005, 2021.

\bibitem[Zhao et~al.(2024{\natexlab{a}})Zhao, Wang, and Liu]{zhao2024bad}
Lingzhe Zhao, Peng Wang, and Peidong Liu.
\newblock Bad-gaussians: Bundle adjusted deblur gaussian splatting.
\newblock \emph{arXiv preprint arXiv:2403.11831}, 2024{\natexlab{a}}.

\bibitem[Zhao et~al.(2022)Zhao, Xiong, Zhao, Yu, Fan, and Huang]{spike_flow}
Rui Zhao, Ruiqin Xiong, Jing Zhao, Zhaofei Yu, Xiaopeng Fan, and Tiejun Huang.
\newblock Learning optical flow from continuous spike streams.
\newblock \emph{NeurIPS}, 35:\penalty0 7905--7920, 2022.

\bibitem[Zhao et~al.(2024{\natexlab{b}})Zhao, Xiong, Zhao, Zhang, Fan, Yu, and Huang]{zhao2024boosting}
Rui Zhao, Ruiqin Xiong, Jing Zhao, Jian Zhang, Xiaopeng Fan, Zhaofei Yu, and Tiejun Huang.
\newblock Boosting spike camera image reconstruction from a perspective of dealing with spike fluctuations.
\newblock In \emph{CVPR}, pages 24955--24965, 2024{\natexlab{b}}.

\bibitem[Zheng et~al.(2021)Zheng, Zheng, Yu, Shi, Tian, and Huang]{STDP_zheng}
Yajing Zheng, Lingxiao Zheng, Zhaofei Yu, Boxin Shi, Yonghong Tian, and Tiejun Huang.
\newblock High-speed image reconstruction through short-term plasticity for spiking cameras.
\newblock In \emph{CVPR}, pages 6358--6367, 2021.

\bibitem[Zheng et~al.(2022)Zheng, Yu, Wang, and Huang]{motion_estimation_zheng}
Yajing Zheng, Zhaofei Yu, Song Wang, and Tiejun Huang.
\newblock Spike-based motion estimation for object tracking through bio-inspired unsupervised learning.
\newblock \emph{IEEE TIP}, 32:\penalty0 335--349, 2022.

\bibitem[Zhou et~al.(2024)Zhou, Lei, Guan, and Yu]{guan3}
Kangrui Zhou, Taihang Lei, Banglei Guan, and Qifeng Yu.
\newblock Event-based depth estimation with dense occlusion.
\newblock \emph{Optics Letters}, 49\penalty0 (12):\penalty0 3376--3379, 2024.

\bibitem[Zhu et~al.(2019)Zhu, Dong, Huang, and Tian]{tfp_tfi}
Lin Zhu, Siwei Dong, Tiejun Huang, and Yonghong Tian.
\newblock A retina-inspired sampling method for visual texture reconstruction.
\newblock In \emph{ICME}, pages 1432--1437. IEEE, 2019.

\bibitem[Zhu et~al.(2020)Zhu, Dong, Li, Huang, and Tian]{zhu2020retina}
Lin Zhu, Siwei Dong, Jianing Li, Tiejun Huang, and Yonghong Tian.
\newblock Retina-like visual image reconstruction via spiking neural model.
\newblock In \emph{CVPR}, pages 1438--1446, 2020.

\bibitem[Zhu et~al.(2024)Zhu, Jia, Zhao, Qi, Wang, and Huang]{zhu2024spikenerf}
Lin Zhu, Kangmin Jia, Yifan Zhao, Yunshan Qi, Lizhi Wang, and Hua Huang.
\newblock Spikenerf: Learning neural radiance fields from continuous spike stream.
\newblock In \emph{Proceedings of the IEEE/CVF Conference on Computer Vision and Pattern Recognition}, pages 6285--6295, 2024.

\end{thebibliography}
